\documentclass[letterpaper, 10 pt, conference]{ieeeconf}  

\IEEEoverridecommandlockouts                              

\usepackage[linesnumbered,ruled,algo2e]{algorithm2e}

\usepackage{times}
\usepackage{epsfig}
\usepackage{graphicx}
\usepackage{amsmath}
\usepackage{amssymb}
\usepackage{commath}
\usepackage{textcomp}

\usepackage{multirow}
\usepackage{array}
\newcolumntype{L}[1]{>{\raggedright\let\newline\\\arraybackslash\hspace{0pt}}m{#1}}
\newcolumntype{C}[1]{>{\centering\let\newline\\\arraybackslash\hspace{0pt}}m{#1}}
\newcolumntype{R}[1]{>{\raggedleft\let\newline\\\arraybackslash\hspace{0pt}}m{#1}}

\usepackage{subfigure}
\usepackage{hyperref}
\newcommand{\RN}[1]{%
  \textup{\uppercase\expandafter{\romannumeral#1}}%
}

\title{\LARGE \bf
PointAtrousGraph: Deep Hierarchical Encoder-Decoder with Point Atrous Convolution for Unorganized 3D Points
}

\author{Liang Pan$^{1}$, Chee-Meng Chew$^{2}$ and Gim Hee Lee$^{3}$
\thanks{$^{1}$Liang Pan is with Advanced Robotics Centre, National University of Singapore 
        {\tt\small pan.liang@u.nus.edu}}%
\thanks{$^{2}$Chee-Meng Chew is with Department of Mechanical Engineering, National University of Singapore 
        {\tt\small chewcm@nus.edu.sg}}%
\thanks{$^{3}$Gim Hee Lee is with Computer Vision and Robotic Perception (CVRP) Lab, Department of Computer Science, National University of Singapore 
        {\tt\small dcslgh@nus.edu.sg}}        
}

\begin{document}

\maketitle
\thispagestyle{empty}
\pagestyle{empty}

\begin{abstract}
    Motivated by the success of encoding multi-scale contextual information for image analysis, we propose our PointAtrousGraph (PAG) - a deep permutation-invariant hierarchical encoder-decoder 
    for efficiently exploiting multi-scale edge features in point clouds. 
    Our PAG is constructed by several novel modules,
    such as Point Atrous Convolution (PAC), Edge-preserved Pooling (EP) and Edge-preserved Unpooling (EU).
    Similar with atrous convolution, our PAC can effectively enlarge receptive fields of filters and thus densely learn multi-scale point features.
    Following the idea of non-overlapping max-pooling operations, we propose our EP to preserve critical edge features during subsampling.
    Correspondingly, 
    our EU 
    modules gradually recover 
    spatial information for edge features.
    In addition, we introduce chained skip subsampling/upsampling modules that directly propagate edge features 
    to the final stage.
    Particularly, our proposed auxiliary loss functions 
    can further improve our performance.
    Experimental results show that our PAG outperform 
    previous state-of-the-art methods on various 3D semantic perception applications.
\end{abstract}
\section{INTRODUCTION}

Owing to the effectiveness in capturing spatially-local correlations of convolution operations, deep convolution neural networks (CNNs) have yielded impressive results for many image-based tasks.
In order to encode multi-scale contextual information, CNNs probe incoming image features with filters or pooling operations at multiple rates and multiple effective fields-of-view~\cite{chen2018encoder}, such as Atrous Spatial Pyramid Pooling (ASPP)~\cite{chen2017rethinking} and Pyramid Pooling Module~\cite{zhao2017pyramid, he2015spatial}.
Both strategies effectively capture the contextual information at multiple scales, 
hence significantly improving the capabilities of CNNs.
Unorganized point cloud is a simple and straight-forward representation of 3D structures, which is frequently applied by modern intelligent robotics applications, such as autonomous driving and human-robot interactions.
However, the unorderedness and irregularity of 3D point clouds make the conventional convolution operation inapplicable.
Despite novel filtering kernels are recently proposed, limited studies have been carried out on designing deep hierarchical encoder-decoder architectures to learn multi-scale 
point
features 
for 3D semantic perception.

\begin{figure}
    \centering
    \vspace{-2mm}
    \includegraphics[width=1\linewidth]{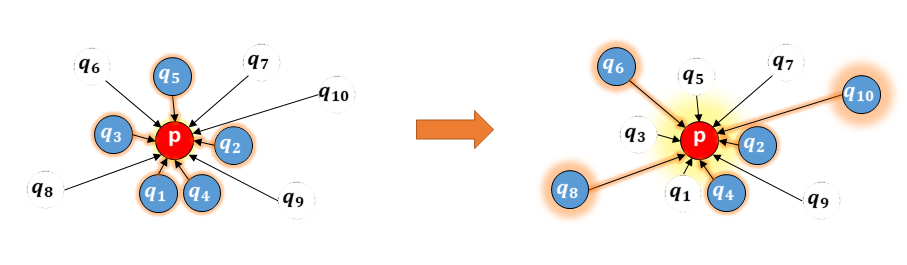} \vspace{-12mm}
    \caption{Point Atrous Convolution (better view in color).  The figure on the left denotes the conventional method in selecting neighboring points \{$q_1, q_2, q_3, q_4, q_5$\}, and the right figure denotes our constructed neighborhood graph \{$q_2, q_4, q_6, q_8, q_{10}$\} with sampling rate equals to 2. By adding the sampling rate parameter, our PAC can perform the convolution operation over a larger field of view without increasing computation load.}
    \vspace{-7mm}
    \label{fig:pac}
\end{figure}

PointNet~\cite{qi2017pointnet} is a pioneering \textbf{\textit{permutation-invariant}} network that directly processes unordered point clouds by using many symmetric functions.
It considers each 3D point independently and meanwhile overlooks local geometric details.
PointNet++~\cite{qi2017pointnet++} adaptively combines multi-scale local features by its proposed abstraction layer.
There are nearly the same encoded local feature vectors if two selected controid points share the same local regions.
These permutation-invariant networks~\cite{dgcnn, xu2018spidercnn} that capture fine geometric structures from local neighborhoods provide better point cloud inference results, which is evident that features of local neighbors can improve deep learning on 3D points.
Nonetheless, receptive fields of their filters are limited to small local regions 
by constructing local neighborhood graphs.
Most recently, PAN~\cite{pan2019pointatrousnet} introduces a novel Point Atrous Convolution (PAC) module, which can effectively enlarge receptive fields of filters
by introducing a sampling rate to equivalently sparsely sample the neighboring point features. 
PAN can densely exploit multi-scale edge features by using PAC modules with different sampling rates.
However, PAN does not have a hierarchical encoder-decoder architecture, which becomes inefficient when dealing with high-dimensional dense point features.

In this paper, we propose the PointAtrousGraph (PAG) - a deep permutation-invariant hierarchical encoder-decoder to exploit multi-scale local geometric details with novel PAC modules for point cloud analysis.
To address the 
\textbf{{overlapped neighborhood graph problem}},
we apply our edge-preserved pooling (EP) operation to preserve critical edge features during subsampling.
Therefore, our PAG can
exploit and preserve multi-scale local geometrical details hierarchically.
In a similar fashion, our edge-preserved unpooling (EU) operation is applied to recover the spatial information of sparse high-dimensional point features.
Furthermore, we introduce tailored chained skip subsampling/upsampling modules to directly propagate point features from each hierarchy.
Additionally, we propose novel auxiliary loss functions,
maximum mean discrepancy (MMD) loss and deeply supervised loss,
which further increases our inference accuracy.
Our PAG also requires less training memory consumption and shorter training time than most existing networks that highly rely on neighborhood graphs for 3D points.
Experiments show that our PAG achieves better performance than previous state-of-the-art methods in various point cloud applications, including 3D object classification, object-part segmentation and 3D semantic segmentation.


\section{RELATED WORK}

\textbf{Unorganized Point Cloud Analysis.}
\textit{Due to the unorderedness of 3D points, a point cloud with $N$ 3D points has a total of $N!$ permutations in the data feeding order.
Hence, it is important for a network to maintain invariance of all possible permutations}~\cite{qi2017pointnet}.
The pioneering work PointNet~\cite{qi2017pointnet} achieves permutation-invariance by applying symmetric functions.
Many following works~\cite{qi2017pointnet++, shen2018mining, dgcnn, xu2018spidercnn} propose more complicated symmetric operations to exploit local geometrical details in 3D points.
Semantic labeling on point cloud is more challenging than classification and object-part segmentation.
SPG~\cite{landrieu2017large} and SGPN~\cite{wang2018sgpn} both construct super point graphs to refine their semantic labeling results.
PAN~\cite{pan2019pointatrousnet} proposes a novel PAC module to effectively exploit multi-scale local edge features.
However, unlike many networks for semantic labeling tasks on images, they~\cite{landrieu2017large, wang2018sgpn, huang2018recurrent, pan2019pointatrousnet} do not have hierarchical encoder-decoder architectures, which limits their performance.

\textbf{Hierarchical Encoder-Decoder.} 
Deep hierarchical encoder-decoder architectures are widely and successfully used for many image-based tasks, such as human pose estimation~\cite{newell2016stacked,pavlakos2017coarse}, semantic segmentation~\cite{ronneberger2015u,badrinarayanan2017segnet,pohlen2017full,lin2017refinenet,peng2017large,amirul2017gated,wojna2017devil,fu2019stacked,zhang2018exfuse,noh2015learning}, optical flow estimation~\cite{dosovitskiy2015flownet,ilg2017flownet}, and object detection~\cite{lin2017feature,shrivastava2016beyond,fu2017dssd}.
The encoder-decoder architecture, stacked hourglass module, is based on the successive steps of pooling and upsampling, which produces impressive results on human pose estimation~\cite{newell2016stacked}.
Lin et al.~\cite{lin2017feature} introduced the feature pyramid network for object detection.
As for semantic segmentation tasks, U-Net~\cite{ronneberger2015u}
and DeconvNet~\cite{noh2015learning} follow the symmetric encoder-decoder architectures, and they refine the segmentation masks by utilizing features in low-level layers.
DeepLabv3+~\cite{chen2018encoder} takes advantage of both the encoder-decoder architecture and the atrous convolution modules to effectively change the fields-of-view of filters to capture multi-scale contextual information, 
which provides new state-of-the-art performance on many semantic segmentation benchmarks.
\textit{Typically, deep hierarchical encoder-decoder architectures contain: (1) \textbf{an encoder module} that progressively reduces the feature resolution, enlarges the receptive fields of filters and captures higher semantic information; (2) \textbf{a decoder module} that gradually recovers the spatial information}~\cite{chen2018encoder}.

\section{METHODS}
Our PointAtrousGraph (PAG) is focused on learning multi-scale edge features by applying a deep hierarchical encoder-decoder architecture.
To maintain the permutation-invariant property, 
our PAG is made up of symmetric functions, such as shared mlp, max-pooling and feature concatenation.
In particular, the PAC module is applied as a fundamental building block, which can effectively learn multi-scale dense edge features in 3D points.
Furthermore, we propose our edge-preserved pooling (EP) operation, which benefits in constructing deep hierarchical networks by preserving critical edge features during subsampling processes.
Our EP operation also enlarges the receptive fields by decreasing 3D point feature density.
In a similar manner, our edge-preserved unpooling (EU) operation gradually recovers the high-dimensional point feature density by considering 3D point spatial locations.
We also directly propagate point features from different hierarchies to the final stage by tailored chained skip subsampling/upsampling modules.
In addition, we propose novel auxiliary losses to further increase our inference accuracy.


\subsection{Point Atrous Convolution for Dense Point Feature}
Two typical methods, ball query and k-nearest neighbors (kNN), are applied 
to exploit local geometric details in point clouds.
However, the ball query algorithm applied by PointNet++~\cite{qi2017pointnet++} always selects the first \#K points in a specified search ball with a predefined radius, which cannot guarantee that closest points can be selected~\cite{komarichev2019cnn}.
In order to exploit sufficient local contextual information in 3D points, conventional networks either construct large neighborhood graphs (large \#K)~\cite{dgcnn, xu2018spidercnn} or concatenate multi-scale local edge features (large \#C$_{f}$)~\cite{jiang2018pointsift, qi2017pointnet++}.
Both above strategies, however, make their networks cumbersome and inefficient.

In contrast, Point Atrous Convolution (PAC)~\cite{pan2019pointatrousnet}
can arbitrarily enlarge its receptive field in dense point features without increasing its computation volume (small \#K and \#C$_{f}$).
Inspired by atrous convolution~\cite{chen2018deeplab} for image analysis, PAC modules introduce an important sampling rate parameter $r$ to equivalently sparsely sample the neighboring point features.
A PAC operation can be formulated as:
\vspace{-2mm}
$$
\vspace{-2mm}
X^{\prime}_p = g(H_{\Theta}(X_p, X_q^{1\cdot r}),\dots,H_{\Theta}(X_p, X_q^{k\cdot r})), \eqno{(1)}
$$
where $X_p$ is the feature of this centroid point $p$, $X_q^{k\cdot r}$ is the feature of the $(k\cdot r)^{th}$ nearest neighbor of point $p$, $r$ is the sampling rate, $k$ is the number of total searched neighboring points, $g(\cdot)$ denotes a max-pooling function, and $H_{\Theta}(\cdot)$ denotes the edge kernel $h_{\theta}\big{(}X_p \oplus (X_p-X_q^{k\cdot r})\big{)}$.
$h_{\theta}$ is a shared mlp layer and $\oplus$ denotes feature concatenation operation.
An example is also illustrated in Fig.~\ref{fig:pac}.



\begin{figure}
    \centering
    \vspace{-0.5mm}
    \includegraphics[width=1\linewidth]{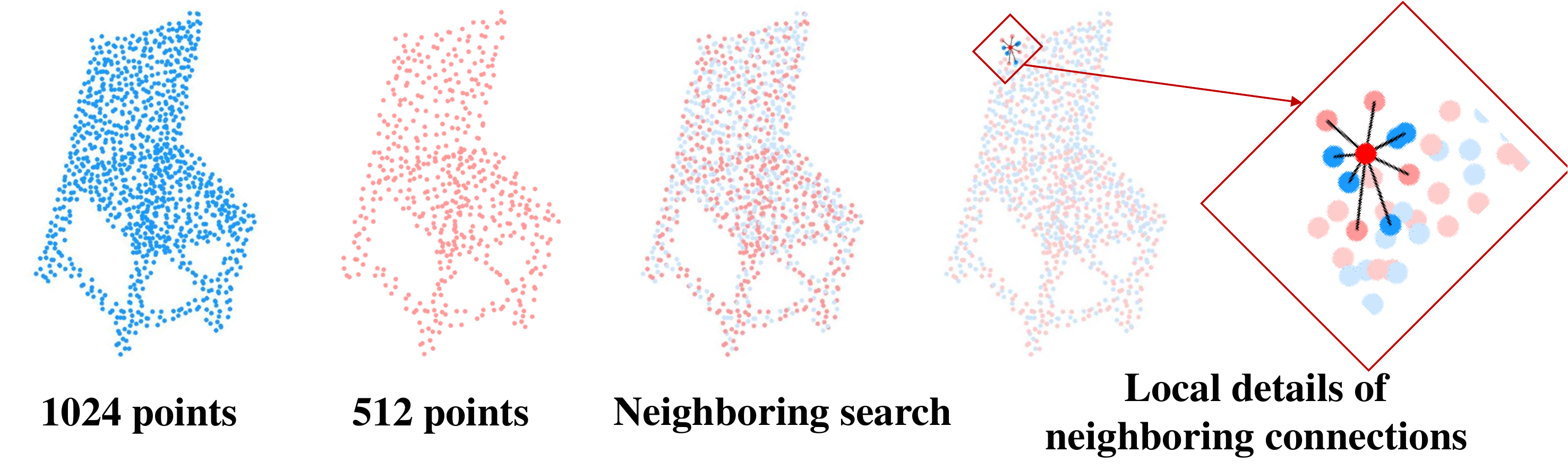}
    \vspace{-9mm}
    \caption{Constructed local neighborhood graphs in 3D points during subsampling processes (better view in color).}
    \vspace{-5mm}
    \label{fig:ep}
\end{figure}

\subsection{Edge-preserved Pooling}
Pooling layers, especially \textbf{\textit{non-overlapping max-pooling}} layers, are widely used in CNNs~\cite{simonyan2014very, he2016deep}, which summarizes the outputs of neighboring groups of neurons in the same kernel map in image domains~\cite{krizhevsky2012imagenet}.
Scherer et al.~\cite{scherer2010evaluation} report that the increment of step size of overlapping pooling windows deteriorates their recognition performance because \textit{maxima in overlapping window regions are merely duplicated in the next layer and neighboring pixels are more correlated.}
Cire\c{s}an et al.~\cite{ciresan2011flexible} replace subsampling layers with non-overlapping max-pooling layers in the CNNs~\cite{lecun1998gradient}, which achieves surprisingly rapid learning speed and better performance.
Various subsampling methods on 3D point clouds~\cite{groh2018flex, qi2017pointnet++} have been proposed.
Nonetheless, they either do not summarize local geometrical features, or ignore the problem caused by overlapped local neighborhood graphs.

\begin{figure}
    \centering
    \includegraphics[width=1\linewidth]{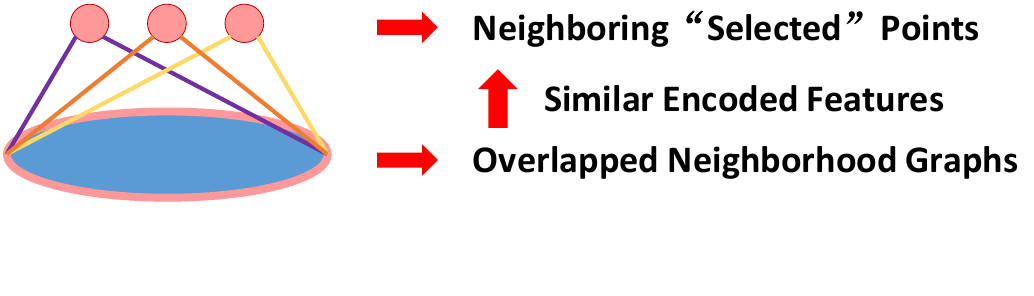}
    \vspace{-15.75mm}
    \caption{The overlapped neighborhood graph problem.}
    \vspace{-6.25mm}
    \label{fig:ongp}
\end{figure}

An example for 3D points subsampling processes is
illustrated in Fig.~\ref{fig:ep}.
The input point feature size is 1,024, and 512 point features are selected after a subsampling process.
If only those features of ``selected'' points are propagated, local geometric details will thus be overlooked.
Another strategy is to propagate edge features by considering the features of local neighboring points (also shown in Fig.~\ref{fig:ep}).
The searched neighboring point features often consist of both ``selected'' point features (shown in pink) and ``discarded'' point features (shown in blue) during subsampling.
Two ``selected'' centroid points can often have overlapped neighborhood graphs, and even share the same local neighboring points.
Hence, if we propagate the encoded neighboring point features, two neighboring ``selected'' points can have similar or even the same features.
The respective information of each 3D point vanishes, especially when multiple subsampling operations are performed to construct a deep hierarchical network for 3D points.
We entitle this as the \textbf{overlapped neighborhood graph problem} shown in Fig.~\ref{fig:ongp}.

In view of this, we propose our edge-preserved pooling (EP) module, which effectively captures local geometrical details while maintaining preserving respective features of each point.
In line with the idea of non-overlapping max-pooling operations, we encode local edge features by considering neighboring point features in the original 1,024 points.
Due to the absence of regular grids in 3D point clouds, we select neighboring point features by constructing neighborhood graphs in metric spaces.
To preserve both distinctive individual point features and local geometrical edge features, our EP module is designed as:
\vspace{-2.5mm}
$$\label{eq:ep}
\vspace{-2.5mm}
\normalsize X^{\prime}_p = X_p \oplus \Large{g} \normalsize (X_q^1,...,X_q^k), \eqno{(2)}
$$
\noindent where $X_p$ is the feature of selected centroid and $X_q^k$ is the point features of its $k$ nearest neighbors.
Consequently, our EP operation explicitly propagates point features of each ``selected'' centroid and also summarizes its local point features, which is consistent with the idea behind non-overlapping max pooling for propagating image features.

\subsection{Edge-preserved Unpooling}
To recover spatial information for image features, low-level image features from encoder are often applied to refine the high-level features in decoder, especially when using a symmetric hierarchical encoder-decoder architecture~\cite{ronneberger2015u, noh2015learning}.
Our edge-preserved unpooling (EU) module also considers the point features of centroids and their local neighboring point features searched in metric spaces.
Unlike PointNet++, our EU module does not need to consider the ``d-dim coordinates'' associated with each point:
\vspace{-2mm}
$$
\vspace{-2mm}
X_p^{\prime} = X_p^e \oplus w(X_q^1,...,X_q^k), \eqno{(3)}
$$
where $X_p^e$ is the corresponding feature propagated from the encoder by a skip connection directly, 
$w(\cdot)$ is the inverse distance weighted average operation and $X_q^k$ is feature of the $k^{th}$ nearest neighbor of point $p$ in its previous hierarchy.

\begin{figure*}
    \centering
    \vspace{-1.5mm}
    \includegraphics[width=1\linewidth]{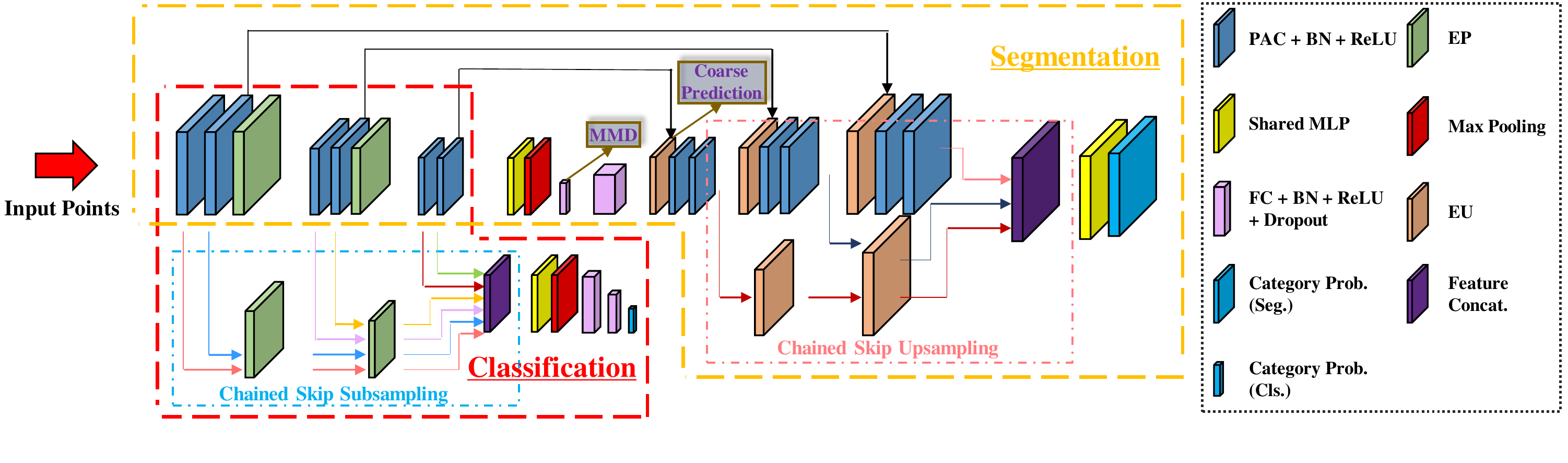}
    \vspace{-13.25mm}
    \caption{Our PointAtrousGraph (PAG) architecture (better view in color). Our classification and segmentation networks have the same designed network encoder architecture. Our classification network is enclosed by red dashed lines, and our segmentation network is enclosed by golden dashed lines. The chained skip subsampling module is applied in our classification network, which is enclosed by blue dash-dotted lines. Likewise, our chained skip upsampling module is enclosed by pink dash-dotted lines in our segmentation network.}
    \vspace{-6mm}
    \label{fig:pag}
\end{figure*}
\subsection{Deep Hierarchical Encoder-Decoder}
Based on our PAC, EP and EU modules, we construct the deep hierarchical encoder-decoder architecture PointAtrousGraph (PAG) to learn multi-scale features for 3D point classification and segmentation tasks (shown in Fig.~\ref{fig:pag}).
The same encoder architecture is applied by our classification and segmentation networks, which consists three hierarchies to gradually reduce the point feature density and meanwhile enlarge the receptive fields for learning higher semantic point features.
Within each hierarchy, we successively lay out two PAC layers with increasing sampling rates to gradually exploit larger local 
geometric details.
Our decoder architectures are designed differently with respect to different applications.
In addition, we also propose different skip connection modules, chained skip subsampling and chained skip upsampling, for classification and segmentation, respectively.

\textbf{Classification Network.}
Our classification network (enclosed by red dashed lines in Fig.~\ref{fig:pag}) aims to encode a global point feature vector by exploiting multi-scale local geometrical details in a point cloud.
The main stream is our network encoder, which unravels the multi-scale contextual information capturing problem by applying many PAC modules in a hierarchical fashion.
In addition to the main stream which consecutively propagates features, we also propose the \textbf{chained skip subsampling} module (enclosed by blue dash-dotted lines in Fig.~\ref{fig:pag}).
The chained skip subsampling module progressively feeds forward features of each hierarchy to the final stage.
In each hierarchy, we select the same set of centroid points with the corresponding EP module to construct local neighborhood graphs.
However, we only propagate features of the neighbors in the chained skip subsampling operations, which is different from the EP module.
After concatenating hierarchical point features from both streams, the global feature is obtained by applying a global max-pooling.
Thereafter, two fully-connected (FC) layers are employed to yield the final classification results. 

\textbf{Segmentation Network.}
A segmentation task can be regarded as a per-point classification.
Hence, our segmentation network (enclosed by golden dashed lines in Fig.~\ref{fig:pag}) densely learns multi-scale edge features for each input 3D point.
Accordingly, we propose a hierarchical decoder to progressively recover the high-dimensional point feature density.
Our decoder for segmentation has a similar architecture as our encoder, which also has three hierarchies.
Likewise, we lay out two PAC layers with decreasing sampling rates to gradually aggregate features for each point.
Particularly, we apply the regressed global features, which largely increases our performance.
We also employ the \textbf{chained skip upsampling} module to consecutively upsample point features of each hierarchy (enclosed by pink dash-dotted lines in Fig.~\ref{fig:pag}).
At each stage, we apply the same upsampling strategy with the EU module with the same constructed local neighborhood graphs.
Unlike EU module, we do not concatenate the centroid features.
Finally, we concatenate all the hierarchical features for each 3D point to perform pre-point predictions.

\subsection{Auxiliary Loss Functions}
We also introduce auxiliary losses, maximum mean discrepancy (MMD) and deeply supervised losses, which are mainly applied for segmentation tasks.

MMD criterion~\cite{zhao2017infovae} that is commonly used in variational auto-encoder architectures is performed over our embeded global point features.
The MMD loss quantifies the similarity between two distributions by comparing all their moments.
By applying the kernel trick, the MMD loss is defined as:
\vspace{-1.5mm}
$$
\vspace{-1.5mm}
L_{mmd}(q\|p) = E_{q(z), q(z\prime)}[k(z, z\prime)] + E_{p(z), p(z\prime)}[k(z, z\prime)] 
$$
$$
\vspace{-2mm}
- 2E_{q(z), p(z\prime)}[k(z, z\prime)],
\eqno{(4)}
$$
where $L_{mmd} \geq 0$, $q(z)$ denotes our embeded feature distribution and $p(z)$ denotes a prior Gaussian distribution (we use $N(\mu=0, \sigma=1.0)$).
$L_{mmd} = 0$ if and only if $q = p$.

Similar with image-based segmentation networks~\cite{zhao2017pyramid, zhao2018psanet}, we add a deeply supervised (cross-entropy) loss $L_{ds}$ at the first stage of our segmentation network decoder.
Along with the master branch (cross-entropy) loss $L_{master}$, auxiliary loss functions also pass through all previous layers.
Consequently, we train our segmentation network by minimizing the following joint loss function:
\vspace{-2mm}
$$
\vspace{-2mm}
L_{all} = L_{master} + w_{mmd}L_{mmd} + w_{ds}L_{ds}, \eqno{(5)}
$$
where $w_{mmd}$ and $w_{ds}$ are designed weights to balance corresponding auxiliary losses.

\subsection{Discussion}
Many networks~\cite{qi2017pointnet, xu2018spidercnn, groh2018flex} that respect the permutation-invariant property are focused on designing convolution kernels for unordered 3D points.
Previous studies~\cite{qi2017pointnet++, dgcnn, xu2018spidercnn, li2018so, shen2018mining} reveal effectiveness of considering local geometric details.
However, limited attentions have been 
received
on designing deep hierarchical encoder-decoder architectures 
for capturing multi-scale local contexts.
In contrast, our PAG 
- a deep hierarchical encoder-decoder, 
adaptively learns multi-scale edge features by varying the field of view for filters in each layer. 
Motivated by the success of exploiting multi-scale contextual information by operations for image feature learning~\cite{badrinarayanan2017segnet, chen2018deeplab, long2015fully, noh2015learning, zhao2017pyramid}, such as atrous convolution, non-overlapping max-pooling and unpooling, we apply novel modules, including PAC, EP and EU, for our multi-scale edge feature learning in unorganized 3D points.
Our PAC module 
effectively enlarges its receptive field in dense point features without increasing training parameters.
With our EP modules,
our PAG is capable of propagating high-dimensional features while preserving local edge features and respective information of each point.
To recover the density for point features, we apply interpolation operations by considering features of those spatial neighboring points.
The proposed auxiliary loss functions further increase our network performance.
Moreover, our hierarchical architecture consists of multiple subsampling operations, which significantly decreases its computation amount and thus largely increases the efficiency.

\section{EXPERIMENTS} \label{sec:exp}
Our PAG is evaluated on three point cloud analysis tasks, including shape classification, object-part segmentation and semantic segmentation.
Without additional processes, our PAG outperform
previous state-of-the-art methods.

\subsection{Implementation Details}
Our networks are implemented with TensorFlow~\cite{abadi2016tensorflow} on an NVIDIA GTX1080Ti GPU.  
We report our results with respect to different input and training strategies to achieve fair comparisons for shape classification.
For segmentation tasks, we follow the same training and evaluation setting in~\cite{qi2017pointnet}.
To improve efficiency, we fix \#K = 10 for all our neighborhood graphs.
We apply the farthest point sampling (FPS) algorithm~\cite{qi2017pointnet++} for subsampling points by considering their metrics (static) or features (dynamic).
Our code and models are publicly available on the project website\href{https://github.com/paul007pl/PointAtrousGraph}{\footnote{https://github.com/paul007pl/PointAtrousGraph}}.

\begin{table}
	\begin{center}
	    \caption{Shape classification results on ModelNet40~\cite{wu20153d}.}
	    \vspace{-1.5mm}
		\begin{tabular}{c|l|c|ccc}
			\hline
			 & \footnotesize{Method} & \footnotesize{Input} & \footnotesize{OA} & \footnotesize{Mem.} & \footnotesize{Time} \\
			\hline
			\multirow{13}*{\rotatebox{90}{\scriptsize{With Up-axis Rotation}}} 
			&\scriptsize{PointNet}~\cite{qi2017pointnet} & \scriptsize{1,024 pts} & \scriptsize{89.2} & \scriptsize{2.4GB} & \scriptsize{3-6h} \\
			&\scriptsize{PointNet++}~\cite{qi2017pointnet++} & \scriptsize{1,024 pts} & \scriptsize{90.7} & \scriptsize{11.1GB} & \scriptsize{$\simeq$20h} \\
			&\scriptsize{Wang(41-spec-cp)}~\cite{wang2018local} & \scriptsize{1,024 pts} & \scriptsize{91.5} & - & \scriptsize{$\simeq$12h} \\
			&\scriptsize{MRTNet(kd-tree)}~\cite{gadelha2018multiresolution} & \scriptsize{4,000 pts} & \scriptsize{91.7} & - & - \\
			&\scriptsize{DGCNN}~\cite{dgcnn} & \scriptsize{1,024 pts} & \scriptsize{92.2} & \scriptsize{8.9GB} & \scriptsize{11.4h} \\
            &\scriptsize{PAN}~\cite{pan2019pointatrousnet} &
            \scriptsize{1,024 pts} & \scriptsize{92.2} &
            \scriptsize{9.2GB} & \scriptsize{8.7h} \\
            &\footnotesize{PAG (ours)} & \scriptsize{1,024 pts} & \footnotesize{\textbf{92.2}} & \scriptsize{2.4GB} & \scriptsize{4.4h}  \\
            \cline{2-6}
			&\scriptsize{PointNet++}~\cite{qi2017pointnet++} & \scriptsize{5,000 pts + n} & \scriptsize{91.9} & - & - \\
			&\scriptsize{SpiderCNN}~\cite{xu2018spidercnn} & \scriptsize{1,024 pts + n} & \scriptsize{92.4} & \scriptsize{11.3GB} & - \\
			&\scriptsize{Wang(41-spec-cp)}~\cite{wang2018local} & \scriptsize{2,048 pts + n} & \scriptsize{92.1} & - & \scriptsize{$\simeq$20h} \\
            &\scriptsize{PAN}~\cite{pan2019pointatrousnet} &
            \scriptsize{1,024 pts + n} & \scriptsize{92.6} &
            \scriptsize{9.2GB} & \scriptsize{8.8h} \\
			&\footnotesize{PAG (ours)} & \scriptsize{1,024 pts + n}& \footnotesize{\textbf{92.7}} & \scriptsize{2.6GB} & \scriptsize{4.5h} \\
			\hline
			\multirow{10}*{\rotatebox{90}{\scriptsize{W/o Up-axis Rotation}}}
			& \scriptsize{KCNet}~\cite{shen2018mining} & \scriptsize{1,024 pts} & \scriptsize{91.0} & - & - \\
			& \scriptsize{Atzmon et al.}~\cite{atzmon2018point} & \scriptsize{1,024 pts} & \scriptsize{92.3} & - & \scriptsize{$\simeq$25h}\\
			&\scriptsize{PointCNN$^{\bullet}$}~\cite{li2018pointcnn} & \scriptsize{1,024 pts} & \scriptsize{92.5} & - & - \\
			&\scriptsize{SO-Net}~\cite{li2018so} & \scriptsize{2,048 pts} & \scriptsize{90.9} & - & \scriptsize{$\simeq$3h} \\
			&\scriptsize{A-CNN}~\cite{komarichev2019cnn} & \scriptsize{1,024 pts} & \scriptsize{92.6} & - & - \\ 
			&\scriptsize{PAN}~\cite{pan2019pointatrousnet} &
            \scriptsize{1,024 pts} & \scriptsize{93.1} &
            \scriptsize{9.2GB} & \scriptsize{8.7h} \\
			&\footnotesize{PAG (ours)} & \scriptsize{1,024 pts} & \footnotesize{\textbf{93.1}} & \scriptsize{2.4GB} & \scriptsize{4.4h} \\
			\cline{2-6}
			&\scriptsize{SO-Net}~\cite{li2018so} & \scriptsize{5,000 pts + n} & \scriptsize{93.4} & - & - \\
			&\scriptsize{PAN}~\cite{pan2019pointatrousnet} &
            \scriptsize{5,000 pts + n} & \scriptsize{93.4} &
            - & - \\
			&\footnotesize{PAG (ours)} & \scriptsize{5,000 pts + n} & \footnotesize{\textbf{93.8}} & - & -\\
			\hline
		\end{tabular}
	\end{center}
	\vspace{-8.75mm}
	\label{table:modelnet40}
\end{table}

\subsection{Shape Classification}
We evaluate the performance of our network for shape classification on the ModelNet40 dataset~\cite{wu20153d}.
ModelNet40 benchmark contains 13,834 CAD models from 40 categories, and it is split into a training (9,843 models) and a test set (2,468 models).
Most 3D models from ModelNet40 are pre-aligned to the common up-direction and horizontal-facing direction.
To better approximate the scenarios in real world applications, the pre-aligned direction can be ignored by applying random up-axis rotations.
For fair comparisons, we report our classification network performance in both training settings, provided Table~\RN{1}.
The column ``OA'' provides the overall classification accuracy (percentage).
The column ``Mem.'' and ``Time'' denotes the required training memory consumption and time, respectively.
According to Table~\RN{1}, 
our PAG yields the best classification accuracy for all the training settings.
Furthermore, our PAG requires smaller memory footprint and shorter time for training than most existing networks.

\textbf{Ablation Study.}
We report the ablation study results in Table~\RN{2}.
Without applying our PAC modules, the classification accuracy dropped due to insufficient contextual information.
Different from previous works~\cite{dgcnn, xu2018spidercnn}, setting the number of selected nearest neighbors as 20 does not increase our classification accuracy.
Furthermore, setting a smaller number of selected nearest neighbors decreases the training memory requirement and also improves the training efficiency.
Additionally, the proposed chained skip subsampling (CSS) module can further improve the classification performance.
In our EP operations, if we only propagate ``centroid'' or ``neighbor'' features, our classification accuracy will drop significantly.

\begin{figure}
    \centering
    \vspace{-10.5mm}
    \includegraphics[width=0.815\linewidth]{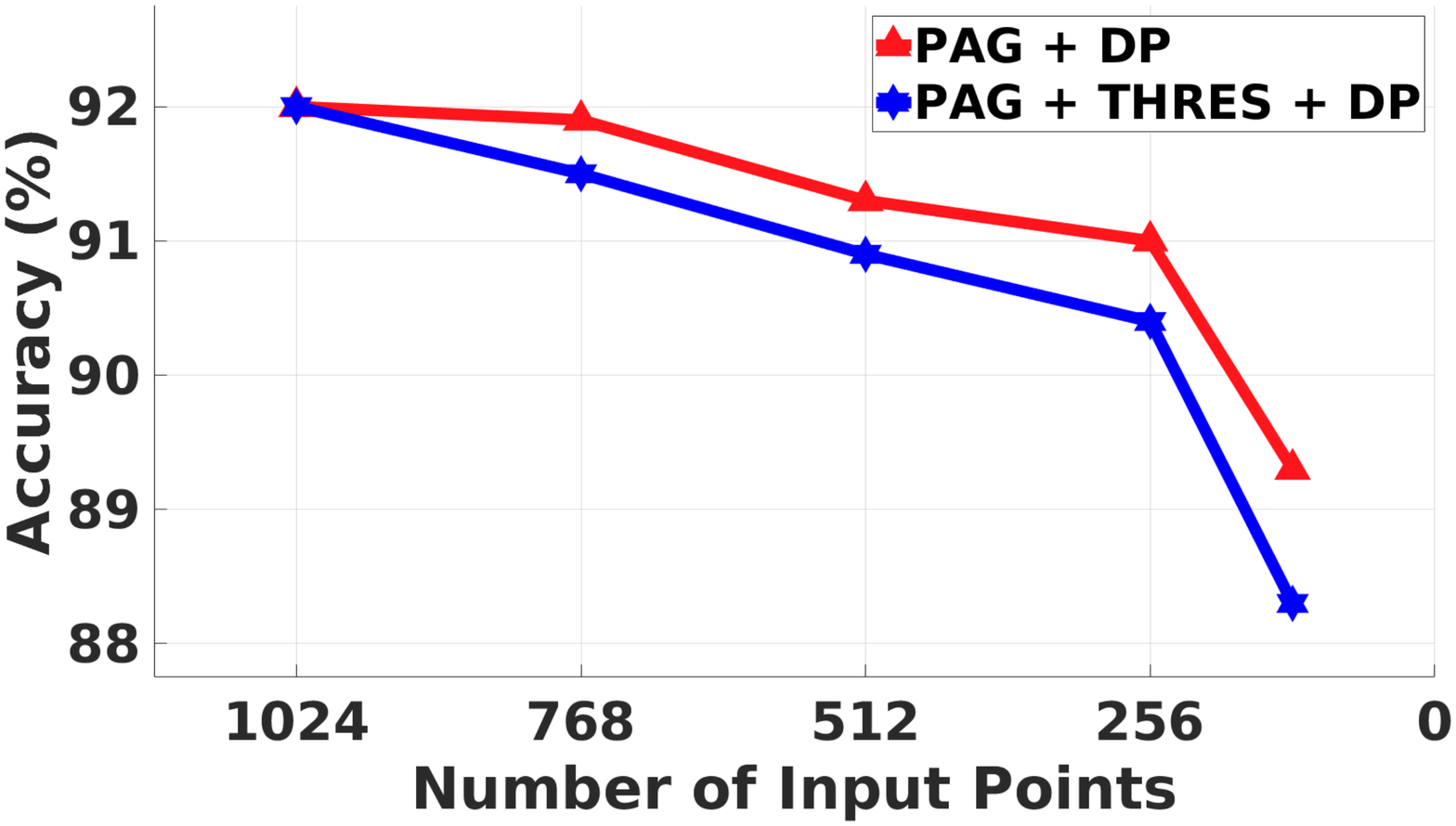}
    \vspace{-11.5mm}
    \caption{Curve showing our advantage of dealing with non-uniform distributed point inputs. DP denotes randomly dropout input points, and THRES denotes additional radii thresholds.}
    \vspace{-1mm}
    \label{fig:dp}
\end{figure}

\begin{table}
    \begin{center}
        \vspace{-1.25mm}
        \caption{Ablation study of the classification network.}
        \vspace{-1.25mm}
        \begin{tabular}{cccc|c}
            \hline
            \scriptsize{K} & \scriptsize{PAC} & \scriptsize{Centroid/Neighbors} & \scriptsize{CSS} & \scriptsize{OA} \\
            \hline
            \scriptsize{10} & \scriptsize{\texttimes} & \scriptsize{both} & \scriptsize{\checkmark} & \scriptsize{92.0} \\
            \scriptsize{20} & \scriptsize{\checkmark} & \scriptsize{both} & \scriptsize{\checkmark} & \scriptsize{91.9} \\
            \scriptsize{10} & \scriptsize{\checkmark} & \scriptsize{both} & \scriptsize{\texttimes} & \scriptsize{92.2} \\
            \scriptsize{10} & \scriptsize{\checkmark} & \scriptsize{centroid} & \scriptsize{\checkmark} & \scriptsize{91.7} \\
            \scriptsize{10} & \scriptsize{\checkmark} & \scriptsize{neighbors} & \scriptsize{\checkmark} & \scriptsize{92.0} \\
            \scriptsize{10} & \scriptsize{\checkmark} & \scriptsize{both} & \scriptsize{\checkmark} & \scriptsize{92.7} \\
            \hline
        \end{tabular}
    \end{center}
    \vspace{-8.25mm}
    \label{table:cls_ablation_study}
\end{table}

\textbf{Robustness to Sampling Density Variation.}
In real applications, point clouds are often partially captured and thus become irregular and incomplete.
Our PAG heavily relies on kNN, which cannot guarantee a fixed scale for exploited geometric details, especially for non-uniformly distributed 3D points.
To this end, we revise the PAC module by applying two searching radii
to bound selected neighboring points in a specified searching field ($r_{min}$, $r_{max}$).
Those searching radii denote the distance between a centroid point and its neighboring point.
In line with~\cite{qi2017pointnet++}, we randomly drop points (DP) to imitate non-uniform and sparse input points.
In our experiments, we observe that applying additional radii negatively impact our performance (shown in Fig.~\ref{fig:dp}).
The reasons can be: 1) our PAG can exploit multi-scale point features to yield robust prediction; 2) it is difficult to manually set the predefined searching thresholds, which cannot be self-adaptive to irregularly incomplete 3D points.


\begin{figure}
    \vspace{-6.5mm}
    \begin{center}
        \includegraphics[width=1\linewidth]{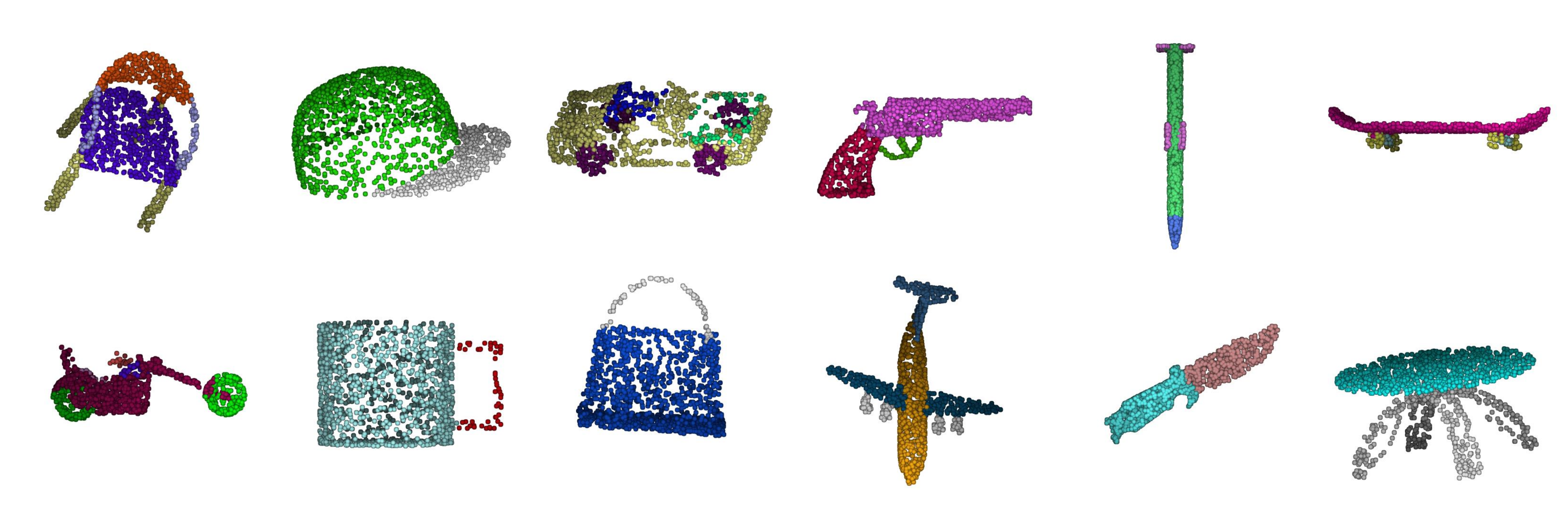}
    \end{center}
    \vspace{-8mm}
    \caption{Qualitative results of the part segmentation task.} 
    \vspace{-0.5mm}
    \label{fig:ps_quality}
\end{figure}

\subsection{Object-part Segmentation}
Object-part segmentation is demonstrated on the ShapeNet-part dataset~\cite{yi2016scalable},
which
contains 16,881 shapes represented as separate point clouds from 16 categories with per-point annotation.
Following previous work~\cite{qi2017pointnet}, 
we split the dataset into a training (14,034 objects) and a test (2,847 objects) set.
We apply the mean Intersection over Union (IoU) metric to evaluate our method on each point.
For each instance, we obtain the mean IoU by averaging IoUs for all part types in the corresponding object category.
The overall instance IoU (``pIoU'') is computed by averaging IoUs over all the tested instances.
Category-wise IoU is computed as the average of all the instances under the corresponding category.
Mean category IoU (``mpIoU'') is thus computed as the mean of all the category-wise IoUs.
In Table~\RN{3},
our model presents the best overall mean IoU at $\textbf{86.4\%}$.
The qualitative segmentation results are shown in Fig.~\ref{fig:ps_quality}.

\textbf{Ablation Study.}
Table~\RN{4} 
provides the results for ablation study on the object-part segmentation task.
The results illustrate that the proposed chained skip upsampling (CSU) module and the extracted global features are beneficial to increasing the segmentation accuracy.
Our proposed PAC module further increases our segmentation accuracy.
Setting the number of selected nearest neighbors as 20 does not increase our segmentation accuracy, but largely increases its training memory consumption.
In our EP operations, ignoring either centroid point features or neighbor point features can lead to a drop of its accuracy.
By adding global point features (GF) and auxiliary loss functions (Aux.L), our segmentation network can yield better results.

\begin{table}
    \vspace{-1.0mm}
    \centering
    \caption{Segmentation comparisons on ShapeNet~\cite{yi2016scalable} and S3DIS~\cite{armeni20163d}.}
    \vspace{-3.5mm}
    \begin{tabular}{l|cc|cc|cc}
        \hline
        \multirow{2}{*}{Method} & \multicolumn{2}{c|}{\scriptsize{ShapeNet}} & \multicolumn{2}{c}{\scriptsize{k-fold} \scriptsize{S3DIS}} & \multicolumn{2}{|c}{\scriptsize{Area-5} \scriptsize{S3DIS}} \\
        \cline{2-7}
         & \scriptsize{mpIoU} & \scriptsize{pIoU} & \scriptsize{OA} & \scriptsize{mIoU} & \scriptsize{OA} & \scriptsize{mIoU} \\
        \hline
        \scriptsize{SGPN$^{\bullet}$}~\cite{wang2018sgpn} & \scriptsize{82.8} & \scriptsize{85.8} & \scriptsize{80.8} & \scriptsize{50.4} & - & - \\
        \scriptsize{RSNet$^{\bullet}$}~\cite{huang2018recurrent} & \scriptsize{81.4} & \scriptsize{84.9} & - & \scriptsize{56.5} & - & - \\
        \scriptsize{SPG$^{\bullet}$}~\cite{landrieu2017large} & - & - & \scriptsize{85.5} & \scriptsize{62.1} & \scriptsize{86.4} & \scriptsize{58.0}\\
        \scriptsize{SegCloud$^{\bullet}$}~\cite{tchapmi2017segcloud} & - & - & - & - & - & \scriptsize{48.9} \\
        \scriptsize{PointCNN$^{\bullet}$}~\cite{li2018pointcnn} & \scriptsize{\textbf{84.6}} & \scriptsize{86.1} & \scriptsize{88.1} & \scriptsize{65.4} & \scriptsize{85.9} & \scriptsize{57.3} \\
        \scriptsize{PCCN}~\cite{wang2018deep} & - & - & - & - & - & \scriptsize{58.3} \\
        \scriptsize{PointNet}~\cite{qi2017pointnet} & \scriptsize{80.4} & \scriptsize{83.7} & \scriptsize{78.5} & \scriptsize{47.6} & - & \scriptsize{41.1}\\
        \scriptsize{PointNet++}~\cite{qi2017pointnet++} & \scriptsize{81.9} & \scriptsize{85.1} & - & - & - & - \\
        \scriptsize{SO-Net}~\cite{li2018so} & \scriptsize{81.0} & \scriptsize{84.9} & - & - & - & - \\
        \scriptsize{SPLATNET$_{3D}$}~\cite{su2018splatnet} & \scriptsize{82.0} & \scriptsize{84.6} & - & - & - & -\\
        \scriptsize{Atzmon et. al}~\cite{atzmon2018point} & \scriptsize{81.8} & \scriptsize{85.1} & - & - & - & - \\
        \scriptsize{SpiderCNN}~\cite{xu2018spidercnn} & \scriptsize{82.4} & \scriptsize{85.3} & - & - & - & - \\
        \scriptsize{DGCNN}~\cite{dgcnn} & \scriptsize{82.3} & \scriptsize{85.2} & \scriptsize{84.1} & \scriptsize{56.1} & - & - \\
        \scriptsize{PAN}~\cite{pan2019pointatrousnet} & \scriptsize{82.6} & \scriptsize{85.7} & \scriptsize{85.9} & \scriptsize{61.4} & - & - \\
        \scriptsize{A-CNN}~\cite{komarichev2019cnn} & \scriptsize{84.0} & \scriptsize{86.1} & \scriptsize{87.3} & \scriptsize{62.9} & - & - \\
        \hline
        \footnotesize{PAG (ours)} & \footnotesize{84.0} & \footnotesize{\textbf{86.4}} & \footnotesize{\textbf{88.1}} & \footnotesize{\textbf{65.9}} & \footnotesize{\textbf{86.8}} & \footnotesize{\textbf{59.3}} \\
        \hline
    \end{tabular}
    \label{tab:seg}
    \vspace{-5mm}
\end{table}

\subsection{Semantic Segmentation}
\begin{figure}
    \vspace{-6.25mm}
    \begin{center}
        \hspace*{-4.5mm}
        \includegraphics[width=1.1\linewidth]{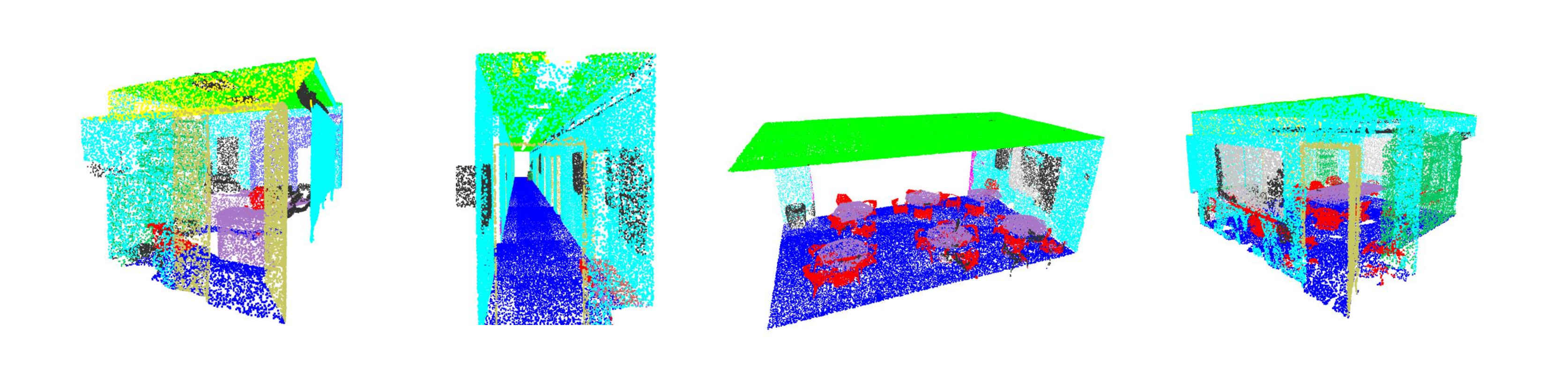}
        \vspace{-10.75mm}
        \caption{Qualitative results on the semantic segmentation task.} 
        \vspace{-4mm}
        \label{fig:ss_quality}
    \end{center}
\end{figure}

We evaluate our model on large-scale 3D semantic segmentation (see Fig.~\ref{fig:ss_quality}) on the Stanford 3D Indoor Semantic Dataset (S3DIS)~\cite{armeni20163d}.
This dataset contains 3D RGB point clouds of 271 rooms from 3 different buildings split into 6 areas.
To achieve fair comparison, we apply the same setting of the training strategy with PointNet~\cite{qi2017pointnet}.
We randomly sample 4,096 points in each block for training, and all the points are used for testing.
The semantic segmentation results (k-fold and Area-5) are provided in Table~\RN{3}. 
Those methods~\cite{li2018pointcnn, wang2018sgpn, huang2018recurrent, landrieu2017large, engelmann2017exploring} (denoted by $^{\bullet}$) obtain unfair advantages by applying specific preprocessing and/or postprocessing procedures, such as super point graphs or recurrent neural networks.
Moreover, they do not maintain the important permutation-invariant property, which makes their results unreliable and sensitive with respect to the points feeding order.
Without additional processes, our permutation-invariant PAG provides the best results on the OA ($\textbf{88.1\%}$ and $\textbf{86.8\%}$) and mIoU ($\textbf{65.9\%}$ and $\textbf{59.3\%}$).

\section{ANALYSIS AND FUTURE DIRECTIONS}
\textbf{Permutation-Invariance Property.}
Our proposed modules are made up of symmetric functions, such as shared mlp, max-pooling and feature concatenation.
Consequently, our PAG is a permutation-invariant network.

\textbf{Sampling Neighboring Points.}
Ball query introduced by PointNet++ is another strategy to sample neighboring points.
However, PointNet++ only select the first \#K points found within the radius, which cannot guarantee that nearest points can be selected.
Furthermore, PointNet++(MSG) search for many neighboring points for 3 times (e.g. \#K=\{16,32,128\}) and then concatenate all exploited features together (e.g. \#C$_f$=64+128+128) in each hierarchy, which makes it computationally expensive~\cite{qi2017pointnet++}.
In contrast, our PAC module equivalently sparsely sample the neighboring points in feature spaces with a sampling rate parameter ``r''.
In our PAG, we consecutively lay out PAC modules in series to aggregate multi-scale edge features without increasing the selected neighboring points (\#K) and the point feature size (\#C$_{f}$).

\textbf{Hierarchical Encoder-Decoder Architecture.}
Inspired by the success of encoding rich contextual image information at multiple scales for image-based applications, we focus on capturing multi-scale local geometrical details in 3D points by a hierarchical encoder-decoder architecture.
Our PAG contains: (1) an encoder module that progressively decreases point feature density, enlarges the field of view of filters and learns higher semantic edge features; (2) a decoder module that gradually recovers the density for high-dimensional point features.
Deep hierarchical CNNs commonly use the non-overlapping max-pooling operations, which decrease the image feature resolution while summarizing local-spatial contextual information.
To resolve the \textbf{overlapped neighborhood graph problem},
we propose our EP operation, which concatenates both ``centroid'' and ``neighbors'' features for hierarchically propagating edge-aware features.
In our ablation studies, we observe that the ``neighbors'' features benefit in generating global features for classification tasks.
The ``centroid'' features are beneficial for propagating respective information of each individual point,
hence leading
to better segmentation results.
Following spatial pyramid pooling operations~\cite{grauman2006pyramid,lazebnik2006beyond}, we can also perform our EP operations at several scales for future studies.

\begin{table}
    \begin{center}
        \caption{Ablation study of the segmentation network.}
        \vspace{-1mm}
        \begin{tabular}{cccccc|c}
            \hline
            \scriptsize{K} & \scriptsize{PAC} & \scriptsize{Centroid/Neighbors} & \scriptsize{CSU} & \scriptsize{GF} & \scriptsize{Aux.L} & \scriptsize{pIoU} \\
            \hline
            \scriptsize{10} & \scriptsize{\checkmark} & \scriptsize{both} & \scriptsize{\checkmark} & \scriptsize{\texttimes} & \scriptsize{\texttimes} & \scriptsize{85.3} \\
            \scriptsize{10} & \scriptsize{\texttimes} & \scriptsize{both} & \scriptsize{\checkmark} & \scriptsize{\checkmark} & \scriptsize{\texttimes} & \scriptsize{85.7} \\
            \scriptsize{20} & \scriptsize{\checkmark} & \scriptsize{both} & \scriptsize{\checkmark} & \scriptsize{\checkmark} & \scriptsize{\texttimes} & \scriptsize{85.7} \\
            \scriptsize{10} & \scriptsize{\checkmark} & \scriptsize{both} & \scriptsize{\texttimes} & \scriptsize{\checkmark} & \scriptsize{\texttimes} & \scriptsize{85.6} \\
            \scriptsize{10} & \scriptsize{\checkmark} & \scriptsize{centroid} & \scriptsize{\checkmark} & \scriptsize{\checkmark} & \scriptsize{\texttimes} & \scriptsize{85.8} \\
            \scriptsize{10} & \scriptsize{\checkmark} & \scriptsize{neighbors} & \scriptsize{\checkmark} & \scriptsize{\checkmark} & \scriptsize{\texttimes} & \scriptsize{85.7} \\
            \scriptsize{10} & \scriptsize{\checkmark} & \scriptsize{both} & \scriptsize{\checkmark} & \scriptsize{\checkmark} & \scriptsize{\texttimes} & \scriptsize{86.1} \\
             \scriptsize{10} & \scriptsize{\checkmark} & \scriptsize{both} & \scriptsize{\checkmark} & \scriptsize{\checkmark} & \scriptsize{\checkmark} & \scriptsize{86.4} \\
            \hline
        \end{tabular}
    \end{center}
    \vspace{-8.0mm}
    \label{table:ps_ablation_study}
\end{table}

\section{CONCLUSION}
We proposed the novel PointAtrousGraph (PAG) architecture to capture multi-scale local geometric details for hierarchical point features learning.
Many novel modules are proposed and then evaluated in our experiments, which significantly increases our evaluation accuracy on classification and segmentation tasks.
In particular, our network consumes much smaller training memory than previous state of the art models.
Hence, our architectures can be conveniently applied for more complicated tasks of large-scale 3D scenes.

\clearpage
\textbf{\large \centering Supplementary Material}

\maketitle

\subsection{Overview}
Our PointAtrousGraph (PAG) is focused on designing a deep permutation-invariant encoder-decoder architecture to learn hierarchical edge features for unorganized point cloud analysis.
Consequently, we emphasize and elaborate those most related and important factors in our main paper: 1) \textbf{unorganized point cloud analysis}; 2) \textbf{permutation-invariance}; 3) \textbf{hierarchical encoder-decoder structure}.
In this documents, more discussions, analysis and qualitative results are provided.

In Sec.~\textit{B}, we discuss and compare other methods and/or representations for organizing 3D scenes.
In Sec.~\textit{C}, we elaborate the advantages and disadvantages for the two categories of deep learning methods for unorganized point cloud analysis: 1) employing symmetric functions; 2) learning canonical orderings.
The details and ablation studies of our network architectures are provided in Sec.~\textit{D},
which also reveals the benefits and effectiveness of 
learning multi-scale local geometric details
with both our PAC and edge-preserved pooling operations.
The Sec.~\textit{E} provides more analysis on our point atrous convolution (PAC).
Thereafter, we illustrate the intuition and effectiveness of applying k-nearest neighbors searches in feature spaces and metric spaces in Sec.~\textit{F}.
In the end, more qualitative results are visualized in Sec.~\textit{G}.

\subsection{Organized 3D Structures vs. Unorganized Point Clouds}\label{sec:org}
Besides the point cloud representation, 3D scenes can also be represented by volumetric grids and scalable indexing structures.

\subsubsection{Volumetric Grid}
Many works~\cite{maturana2015voxnet, qi2016volumetric, wu20153d, zhou2017voxelnet, le2018pointgrid} use volumetric grids to regularize point clouds for deep learning.
VoxNet~\cite{maturana2015voxnet} applies 3D CNN over a volumetric occupancy grid, but expensive computational and memory requirements limit its performance.
VoxelNet~\cite{zhou2017voxelnet} encodes a point cloud as a descriptive volumetric representation,  and then it is connected with a region proposal network to generate detections.
PointGrid~\cite{le2018pointgrid} incorporates the 3D points within each volumetric grid cell. 
This allows the network to learn higher order local features of a 3D point cloud with a smaller memory footprint.
SSCN~\cite{graham20173d} introduces sub-manifold sparse convolutions that maintain the sparsity of the voxelized input over many layers by fixing the location of active sites.

\subsubsection{Scalable Indexing Structure}
Scalable indexing structures, e.g. kd-tree and octree, are used in many existing works~\cite{riegler2017octnet, wang2017cnn, klokov2017escape}. 
OctNet~\cite{riegler2017octnet} hierarchically partitions the 3D space into a set of unbalanced octrees, which leads to a significant reduction in computational and memory resources.
O-CNN~\cite{wang2017cnn} stores the octant information and CNN features into the graphics memory by designing an octree data structure.
3D CNN operations are only performed on those octants occupied by the 3D shape surfaces.
Its computation time and memory requirement grow quadratically with the depth of the designed octree.
Kd-Net~\cite{klokov2017escape} uses a kd-tree structure to construct the computational graph with shared learnable parameters.
Similar with CNNs, Kd-Net computes a sequence of hierarchical representations in a feed-forward bottom-up fashion.

\subsubsection{Comparison and Discussion}
\paragraph{Comparison with the Volumetric Grid Representation.}
The main advantages of using volumetric grids are as follows:
1) volumetric grid is the most straight-forward and understandable method to organize irregular 3D observations;
2) volumetric grids are regular and uniformly distributed;
3) a occupation grid or a distance field can be used as a mean of data representation with the help of a volumetric grid.

However, its disadvantages are also apparent:
1) volumetric grid is notorious for its memory and computationally inefficiency;
2) volumetric grid does not scale well for dense 3D data, which is incapable of exploiting the rich and detailed geometry of original 3D data;
3) the inference results can be sensitive to the size of the volumetric grid as well as the 3D grid resolution.

\paragraph{Comparison with the Scalable Indexing Structure Representation.}
Owing to the inefficiency of organizing 3D points with volumetric grids, scalable indexing structures, such as kd-tree and octree, are utilized to organize 3D point clouds.
It forms the computational graph to apply 3D convolutions level by level, where the convolution kernels often share learnable parameters.
Consequently, the sequence of hierarchical representations can be computed in a feedforward bottom-up fashion~\cite{le2018pointgrid}.
The most important advantage on applying these structures is their capability on exploiting the sparsity of 3D data, and thus adaptively allocate computational and memory resources with respect to the data density.
However, these data structures are more complicated, which makes it a very difficult task to implement the networks.

\paragraph{Discussion.}
Point cloud is the most straight-forward representation and simple representation of 3D structures.
Designing networks directly on 3D points can be much more flexible and convenient in comparison with applying the two organization methods above.
Furthermore, observed 3D scenes in real applications, are usually irregularly and unpredictably distributed and occluded, which can lead to very unreliable understandings if they are organized by regular structures.
The major difficult of applying the point cloud representation is the absence of 3D point ordering as well as the irregular distribution.
To mitigate this limitation, two categories of networks are proposed.
The methods belonging to the first category are following the pioneering network, PointNet~\cite{qi2017pointnet}, which applies symmetric functions to maintain the important permutation-invariant property.
The other category is trying to learn a canonical ordering in 3D points, and the representative network is PointCNN~\cite{li2018pointcnn}.
We are going to discuss and compare the two categories of methods in detail in the next section.

\begin{table*}
    \centering
    \caption{Different Network Architectures and Evaluated Accuracy.}
    \vspace{-3mm}
    \begin{tabular}{c|ccc|c}
        \hline
        Network Configuration & Hierarchy & K & Subsampling Rate & Acc. \\
        \hline
        $([32, 1], \quad [64, 2], \quad [128, 2], \quad [256, 4], \quad [512, 8])$ & 1 & 10 & None & 92.4 \\
        $([32, 1], \quad [64, 2], \quad [128, 4], \quad [256, 8])$ & 1 & 10 & None & 92.3 \\
        $([32, 2]; \quad [64, 2]; \quad [128, 2]; \quad [256, 2]; \quad [512, 2])$ & 5 & 10 & 2 & 92.2 \\
        $([32, 2]; \quad [64, 2]; \quad [128, 2]; \quad [256, 2]; \quad [512, 2])$ & 5 & 5 & 2 & 92.1 \\
        $([32, 1]; \quad [64, 1]; \quad [128, 2]; \quad [256, 2]; \quad [512, 2])$ & 5 & 10 & 2 & 92.2 \\
        $([32, 1], [32, 2]; \quad [64, 1], [64, 2]; \quad [128, 1], [128, 2]; \quad [256, 1], [256, 2])$ & 4 & 10 & 2 & 92.3 \\
        $([32, 1], [64, 2]; \quad [128, 1], [256, 2])$ & 2 & 10 & 2 & 92.3 \\
        $([64, 1], [64, 2]; \quad [128, 1], [128, 2])$ & 2 & 10 & 2 & 92.1 \\
        $([64, 1], [64, 2]; \quad [256, 1], [256, 2])$ & 2 & 10 & 2 & 92.4 \\
        $([64, 1], [64, 2]; \quad [256, 1], [256, 2])$ & 2 & 10 & 4 & 92.3 \\
        $([32, 1], [32, 2]; \quad [64, 1], [64, 2]; \quad [128, 1], [128, 2])$ & 3 & 10 & 2 & 92.1 \\
        $([64, 2], [64, 4]; \quad [128, 2], [128, 4]; \quad [256, 2], [256, 4])$ & 3 & 5 & 2 & 92.0 \\
        $([64, 1], [64, 2]; \quad [128, 1], [128, 2]; \quad [256, 1], [256, 2])$ & 3 & 10 & 2 & 92.3 \\
        $([32, 1], [32, 2]; \quad [64, 1], [64, 2]; \quad [128, 1], [128, 2])$ & 3 & 5 & 4 & 91.5 \\
        $([64, 2], [64, 4]; \quad [128, 2], [128, 4]; \quad [256, 2], [256, 4])$ & 3 & 10 & 4 & 91.8 \\
        $([64, 1], [64, 2]; \quad [256, 1], [256, 2]; \quad [512, 1], [512, 2])$ & 3 & 10 & 4 & 91.7 \\
        $([64, 1], [64, 2]; \quad [128, 1], [128, 2]; \quad [256, 1], [256, 2])$ & 3 & 10 & 4 & 92.7 \\
    \end{tabular}
    \vspace{-6mm}
    \label{table:atrou_valid}
\end{table*}

\subsection{Learning Canonical Ordering vs. Utilizing Symmetric Functions}\label{sec:PI}
Due to the unorderedness of 3D points, a point cloud with N 3D points has a total of $N!$ permutations in the data feeding order.
We hope that our inference results can be stable, reliable and also invariant with the order for input points.
Two strategies are thus proposed:
1) preserving the permutation-invariance by only using symmetric functions;
2) learning a canonical ordering.
PointNet~\cite{qi2017pointnet} and its following works~\cite{qi2017pointnet++, xu2018spidercnn, dgcnn, shen2018mining, li2018so} maintain the permutation-invariant property by only applying symmetric functions, which is effectively applied and validated by many applications.
Belonging to this category, all the operations in our PAG are symmetric, which makes our PAG a permutation-invariant architecture.

PointCNN~\cite{li2018pointcnn} is recently released, which aims to learn an $\chi$-transformation from the input points to permutate themselves into canonical orders.
Ideally, their inference results can be invariant to the input point ordering with the help of this $\chi$-transformation.
\textit{\textbf{However, their learned $\chi$-transformations are far from ideal, especially with the permutation equivalence aspect.}}
In other words, they did not manage to resolve the canonical ordering learning problem with their $\chi$-transformations.
Although they provided state-of-the-art results on various benchmarks, their results can be dependent and sensitive to the 3D points feeding orders.
Furthermore, they do not provide the covariance analysis, which is important and necessary for analyzing the influence of the point feeding orderings on the inference accuracy.
In addition, PointCNN prepare their own datasets for training and testing, and they also apply many specific strategies during training.
All these operations provide additional and unfair advantages for PointCNN.
Consequently, the results provided by PointCNN cannot be directly compared with those permutation-invariant networks that follow the idea behind PointNet.
As an early-stage work, a rigorous understanding of PointCNN is still an open and unsolved problem.

\subsection{Network Architecture Details}\label{sec:struc}
Our classification and segmentation networks share the same encoder architecture.
The encoder configuration that provides the best results is Encoder([64, 1], [64, 2]; [128, 1], [128, 2]; [256, 1], [256, 2]), where $[C, r]$ denotes the feature channel size $C$ and sampling rate $r$, respectively.
The subsampling rate between each two hierarchies is set as 4.
For example, if the input has 1,024 points, then 256 points will be selected after an EP process.

After the encoder, we add a shared mlp layer to project each point feature to high-dimensional feature space.
The fully-connected (FC) layers in our classification network have the following feature size: $512$, $256$ and the number of total classes in the evaluated dataset.
As for our segmentation network, we set the FC layers with the following feature size: $512$ and $1024$.
The decoder of our segmentation network has a similar architecture with our encoder, but all the configurations are set inversely: Decoder([256, 2], [256, 1]; [128, 2], [128, 1]; [64, 2], [64, 1]).
Likewise, the upsampling rate is also set as 4.
Each layer in the chained skip subsampling/upsampling layers has the same configuration with the corresponding layer in the main stream.

The configuration of our decoder for classification network is selected and fixed in advance, and our decoder for segmentation network follows the setting of our network encoder.
Therefore, it is important to find the best configuration for our network encoder.
We have evaluated multiple different encoder architectures on the classification task, as reported
in Table~\RN{5}.
In our experiments, we observe that the idea of learning edge features at different scales of receptive fields indeed improves our inference accuracy as well as convergence speed.
Only apply the PAC modules (hierarchy equals to 1) can improve the inference accuracy, but the training can be inefficient, especially when dealing with dense high-dimensional point features.
Applying hierarchical architectures without the PAC modules can lead to better efficiency but limited receptive fields in each network hierarchy.
Therefore, our PAC and EP modules benefit each other for improving the learning capabilities on exploiting multi-scale local geometric details.
\textbf{On the one hand, PAC can enlarge the field-of-view for filters without decreasing point feature density.
On the other hand, EP increases the receptive fields as well as the network efficiency.}
There can be some fluctuations in training performance from time to time. 
However, our selected configuration often provides relatively steady results, especially when applying the non-rotation training strategy.

\textbf{Time and Memory Complexity.}
Table~\RN{6} reports the total number of parameters that need to be trained, and the average inference time of a few representative networks.
Compared with the other networks~\cite{dgcnn, xu2018spidercnn} that highly rely on nearest neighborhood graphs, our model takes shorter time for a single forward process.

\begin{table}
    \begin{center}
        \caption{Model size and inference time. }
        \vspace{-1mm}
        \begin{tabular}{l|cc}
            \hline
            \small{Method} & \small{\#Params(M)} & \small{Fwd.(ms)} \\
            \hline
            \footnotesize{PointNet(vanilla)}~\cite{qi2017pointnet} & \footnotesize{0.8} & \footnotesize{11.5} \\
            \footnotesize{PointNet}~\cite{qi2017pointnet} & \footnotesize{3.5} & \footnotesize{29.3} \\
            \footnotesize{PointNet++(msg)}~\cite{qi2017pointnet++} & \footnotesize{1.7} & \footnotesize{98.1} \\
            \footnotesize{DGCNN}~\cite{dgcnn} & \footnotesize{1.8} & \footnotesize{173.7}\\
            \footnotesize{SpiderCNN(3-layers)}~\cite{xu2018spidercnn} & \footnotesize{3.2} & \footnotesize{275.4} \\
            \hline
            \footnotesize{PAG (Ours)} & \footnotesize{1.8} & \footnotesize{109.8} \\
            \hline
        \end{tabular}
    \end{center}
    \vspace{-8mm}
    \label{table:model size and inference time}
\end{table}

\subsection{Point Atrous Convolution Analysis}\label{sec:atrous}
Our point atrous convolution is inspired by the atrous convolution~\cite{chen2018deeplab} in image domain.
Conventional convolution is restricted in local regions due to applying the $3\times3$ convolution kernels.
To enlarge the receptive fields of convolution kernels without increasing the overall computation amount, a sampling rate $r$ is introduced in conventional convolution kernel for selecting and sampling the neighboring image features.
The most important advantage of atrous convolution is that it effectively enlarges the field of view for each operation kernel without sacrificing the feature density, which is evaluated by many image-based applications, especially the semantic segmentation tasks~\cite{chen2018deeplab, chen2017rethinking, chen2018encoder, yu2015multi}.

In view of the success of atrous convolution in image domains, we extend the idea of atrous convolution for point cloud analysis.
In a similar spirit, we apply the sampling rate $r$ to equivalently sparsely sample and select neighboring points to exploit and aggregate contextual information from a larger local neighborhood region.
Hence, we introduce our point atrous convolution (PAC).
Our PAC enables our network to exploit local contextual information in 3D points with flexible scales.
However, it can be inefficient when we propagate point features to the high-dimensional feature space.
In this work, we also propose our edge-preserved pooling (EP) operation, which is also able to enlarge the receptive fields for our convolution kernels.
Nonetheless, the point feature density is decreased after an EP operation.
In our experiments, \textbf{we observe that both operations benefit each other mutually.}
More results with different network architectures are further elaborated in Sec.~\ref{sec:struc}.

\subsection{Mixture of K-Nearest Neighbors Search in Feature Spaces and Metric Spaces}\label{sec:knn}
\begin{figure}
    \centering
    \includegraphics[width=1\linewidth]{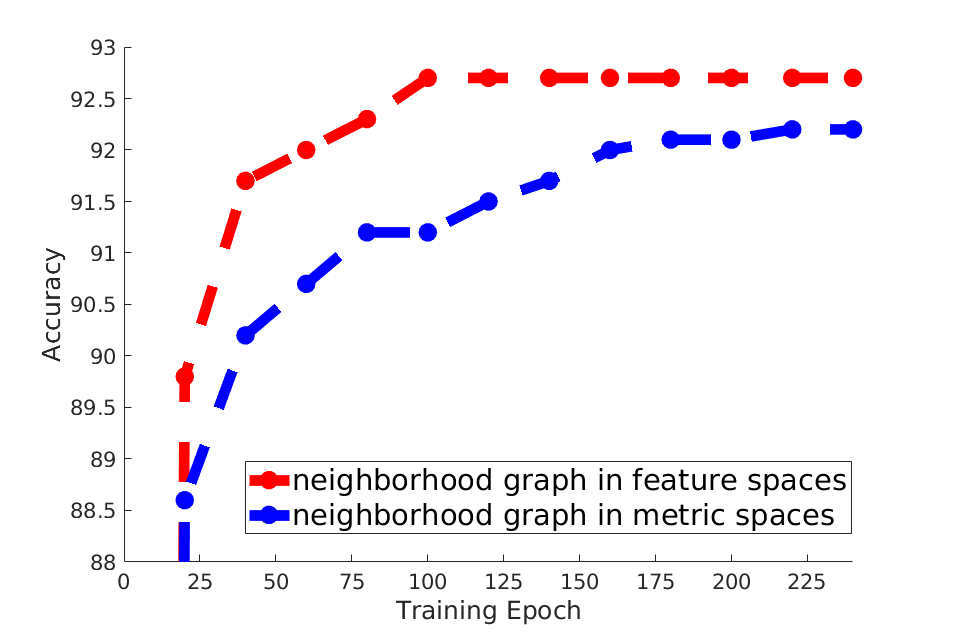}
    \vspace{-7mm}
    \caption{Test accuracy (show the best) for PAC selecting neighboring points in feature spaces and metric spaces during training.}
    \vspace{-5mm}
    \label{fig:cvg}
\end{figure}

In our PAG architecture, we exploit neighboring point features in feature spaces and metric spaces.
Our PAC module is applied to learn edge features by considering local neighboring point features that are selected in feature spaces.
In our experiments, we observe that constructing neighborhood graphs in feature spaces for edge feature extracting can lead to better inference results as well as faster convergence speed during training than constructing neighborhood graphs in metric spaces, as shown in Fig.~\ref{fig:cvg}.

However, as for constructing deep hierarchical networks, the proposed edge-preserved pooling (EP) and edge-preserved unpooling (EU) both select neighboring points in the metric spaces.
The intuitions of designing EP and EU modules are from the pooling and unpooling operations in image domains.
Those operations for image analysis summarize/recover local contextual information with the help of regular image patterns.
Due to the absence of regular grids in 3D points, the local geometric details can be summarized/recovered by constructing local neighborhood graphs in metric spaces instead.
Moreover, the constructed neighborhood graphs in metric spaces are static, whereas the neighborhood graphs in feature spaces are dynamic.
Accordingly, maintaining a stable hierarchical structure can benefit the convergence of the overall network structure, especially the EU operation, which is the second reason.
Last but not the least, we discuss and analyze an important problem - the overlapped neighborhood graph problem, in the main paper.
If we select neighboring point features in feature spaces, the overlapped neighborhood graph problem can be frequently encountered, which also deviates from our original intention - to capture characteristic local geometric details.

\subsection{More Qualitative Results}\label{sec:results}
More qualitative results are visualized in this section.
\begin{figure}[h!]
    \centering
    \vspace{-4mm}
    \hspace*{-4mm}
    \includegraphics[width=1\linewidth]{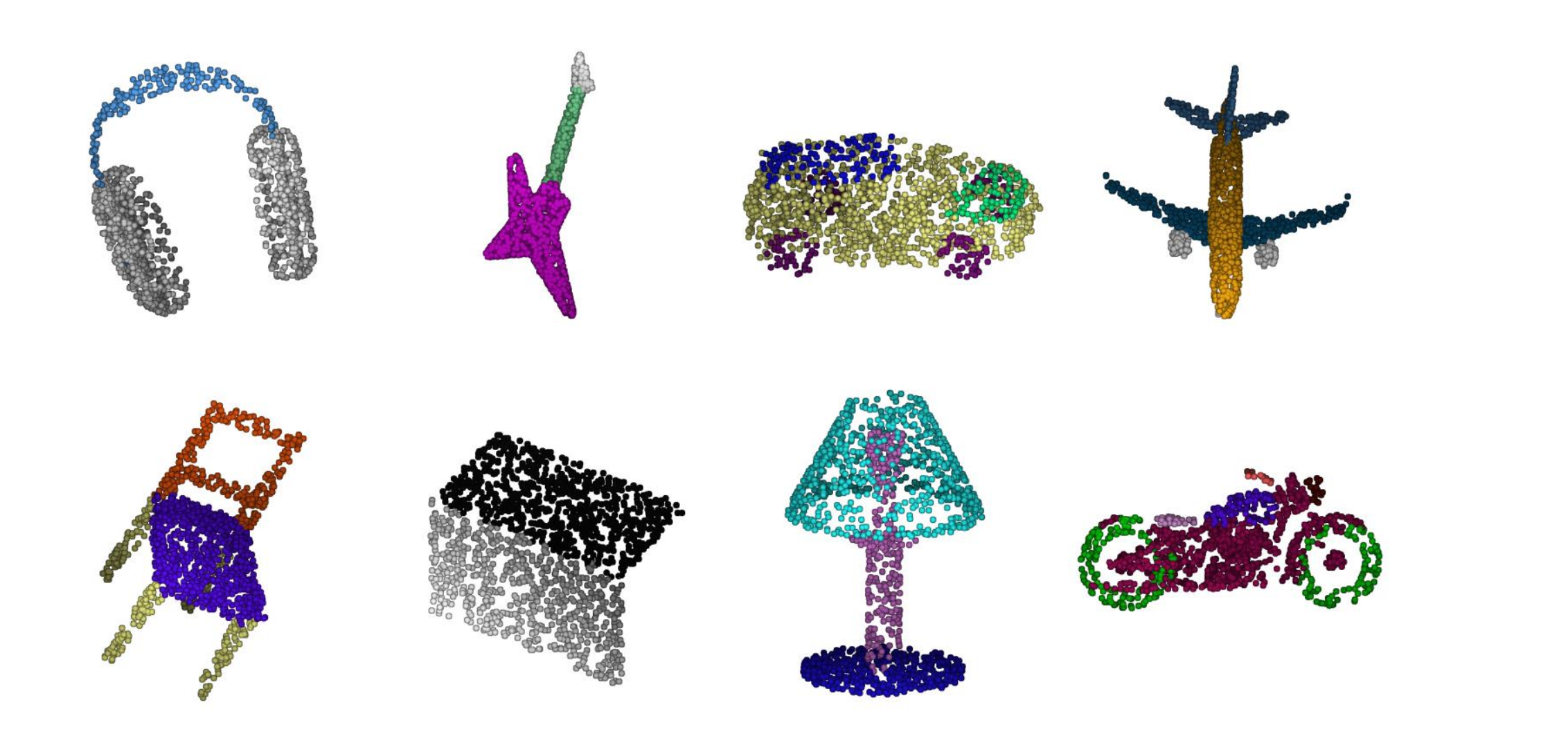}
    \vspace{-4mm}
    \caption{More qualitative results on ShapeNet~\cite{yi2016scalable}}
    \label{fig:my_label}
\end{figure}

\begin{figure}
    \centering
    \vspace{-3mm}
    \subfigure{
        \hspace*{-12mm}
        \includegraphics[width=0.8\linewidth]{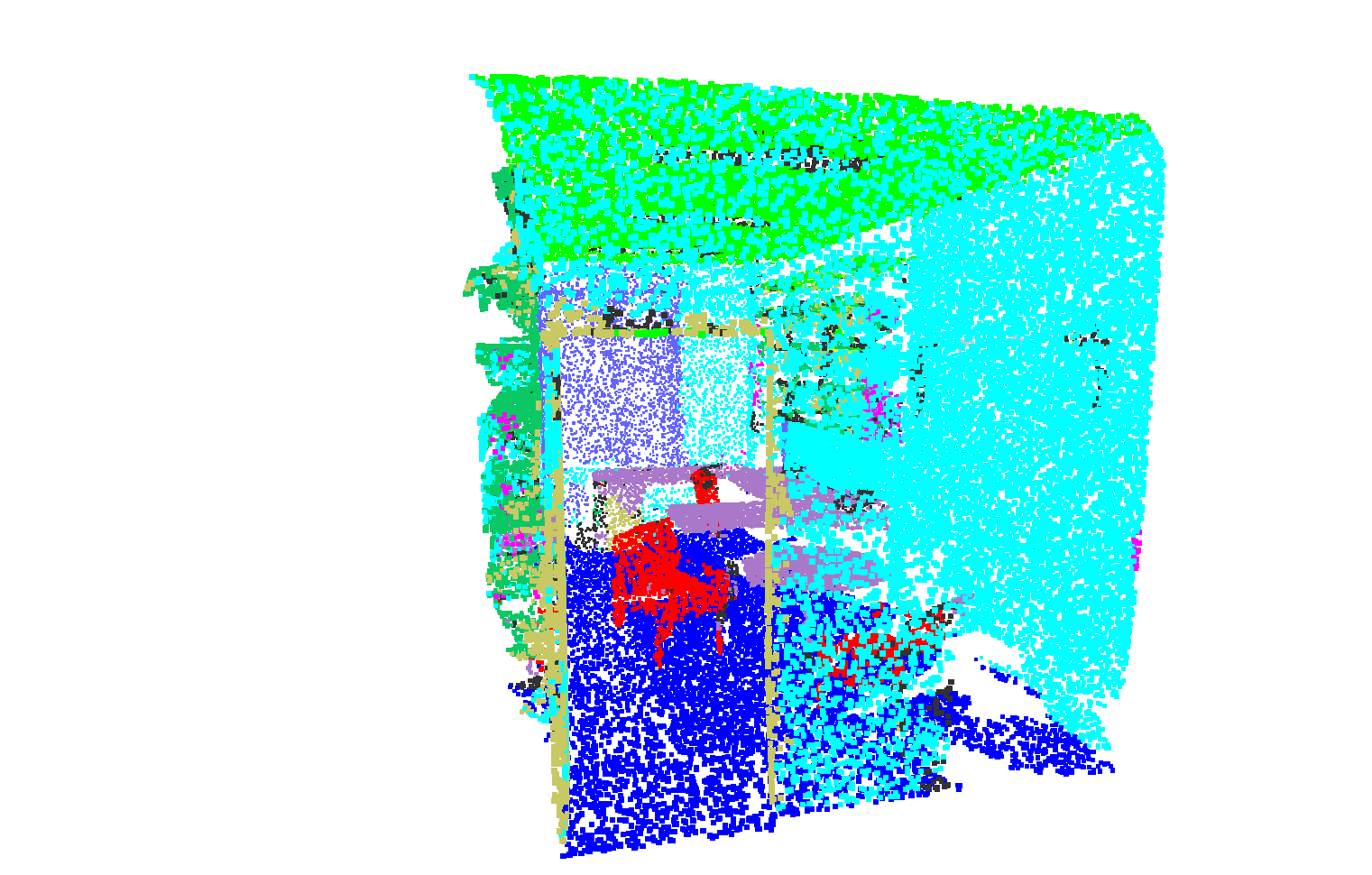}
    }
    \vspace{-5mm}
    \subfigure{
        \hspace*{1mm}
        \includegraphics[width=0.8\linewidth]{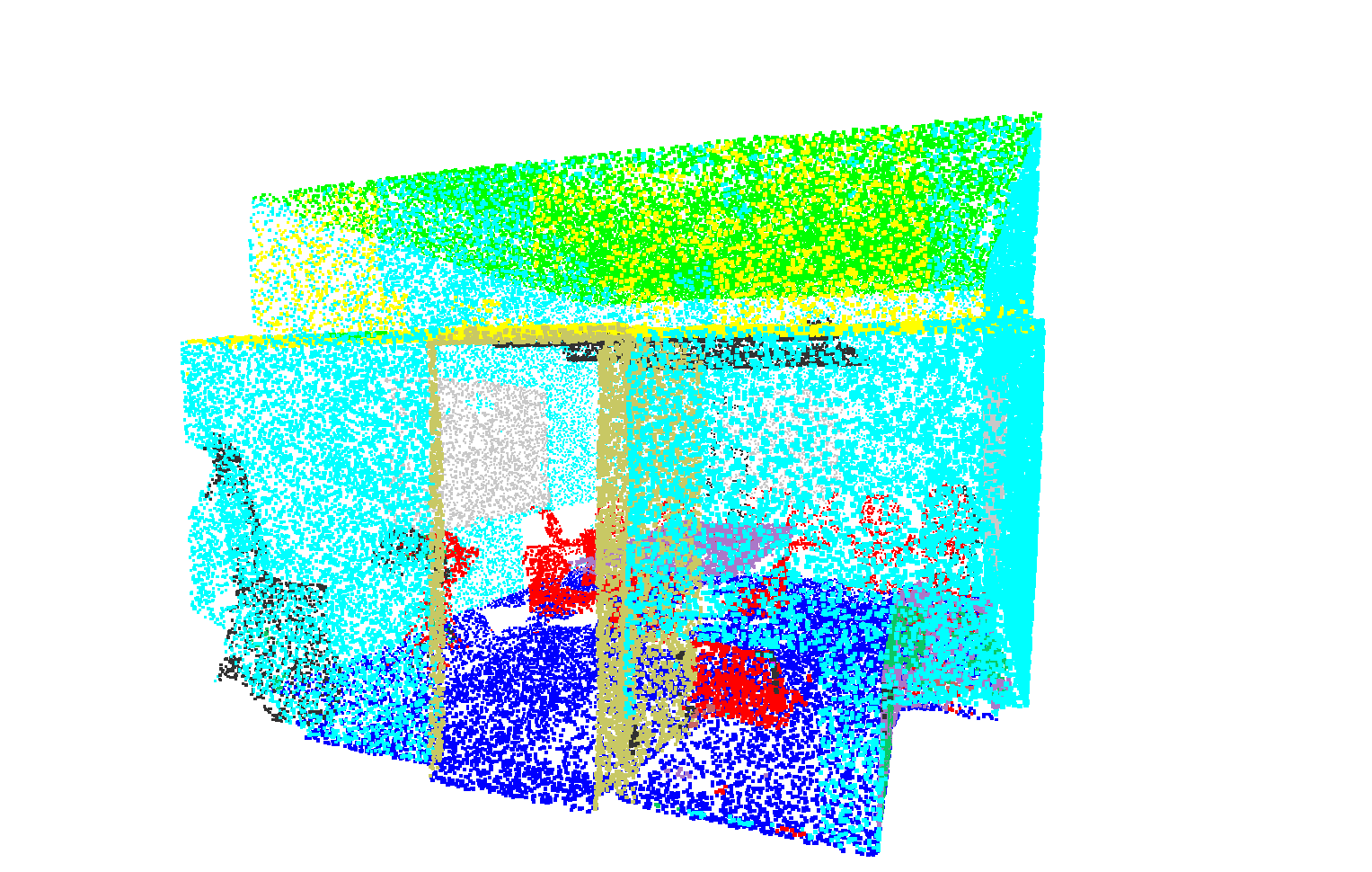}
    }
    \vspace{-6mm}
    \subfigure{
        \includegraphics[width=0.8\linewidth]{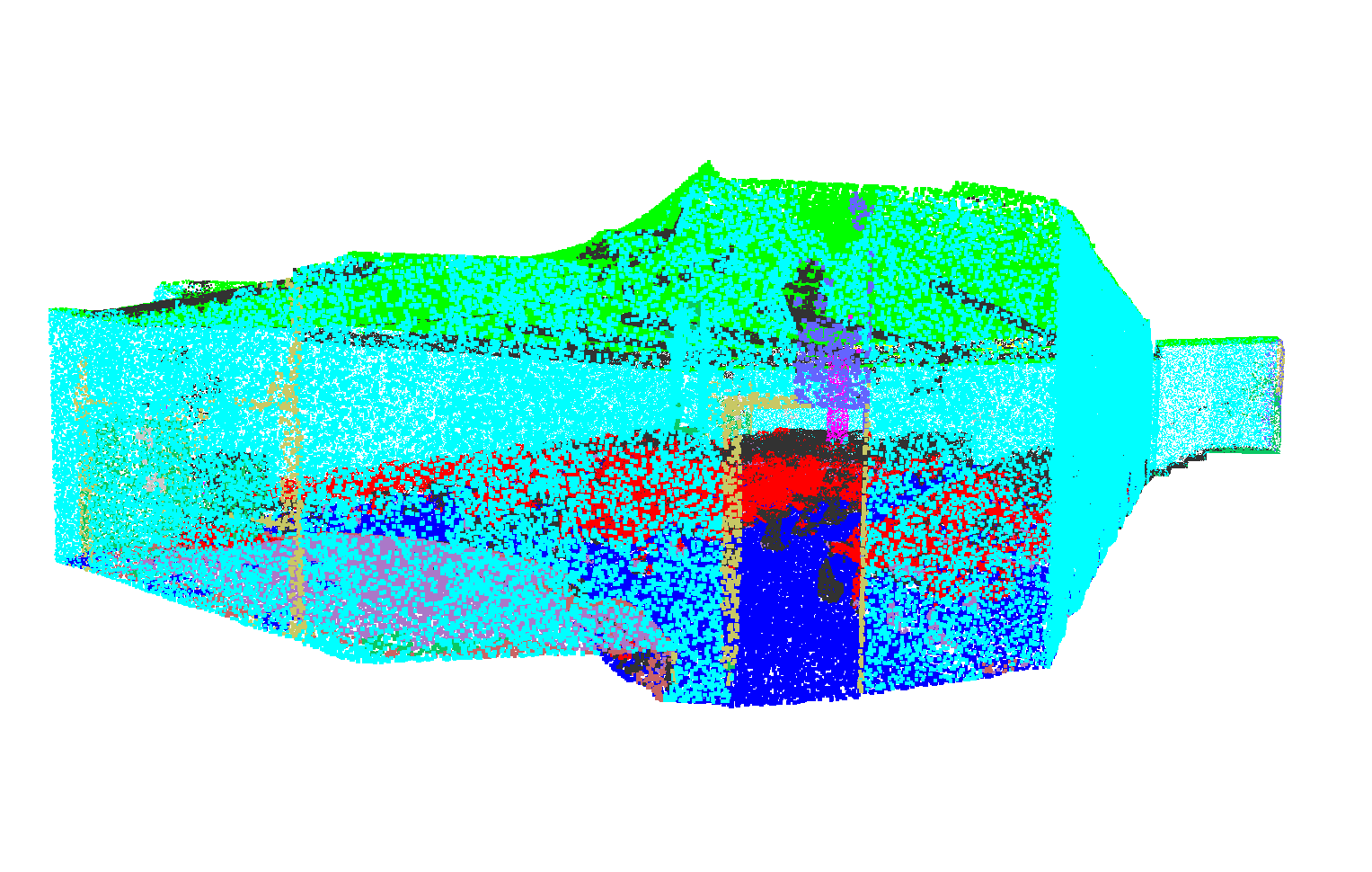}
    }
    \vspace{-8mm}
    \subfigure{
        \includegraphics[width=0.8\linewidth]{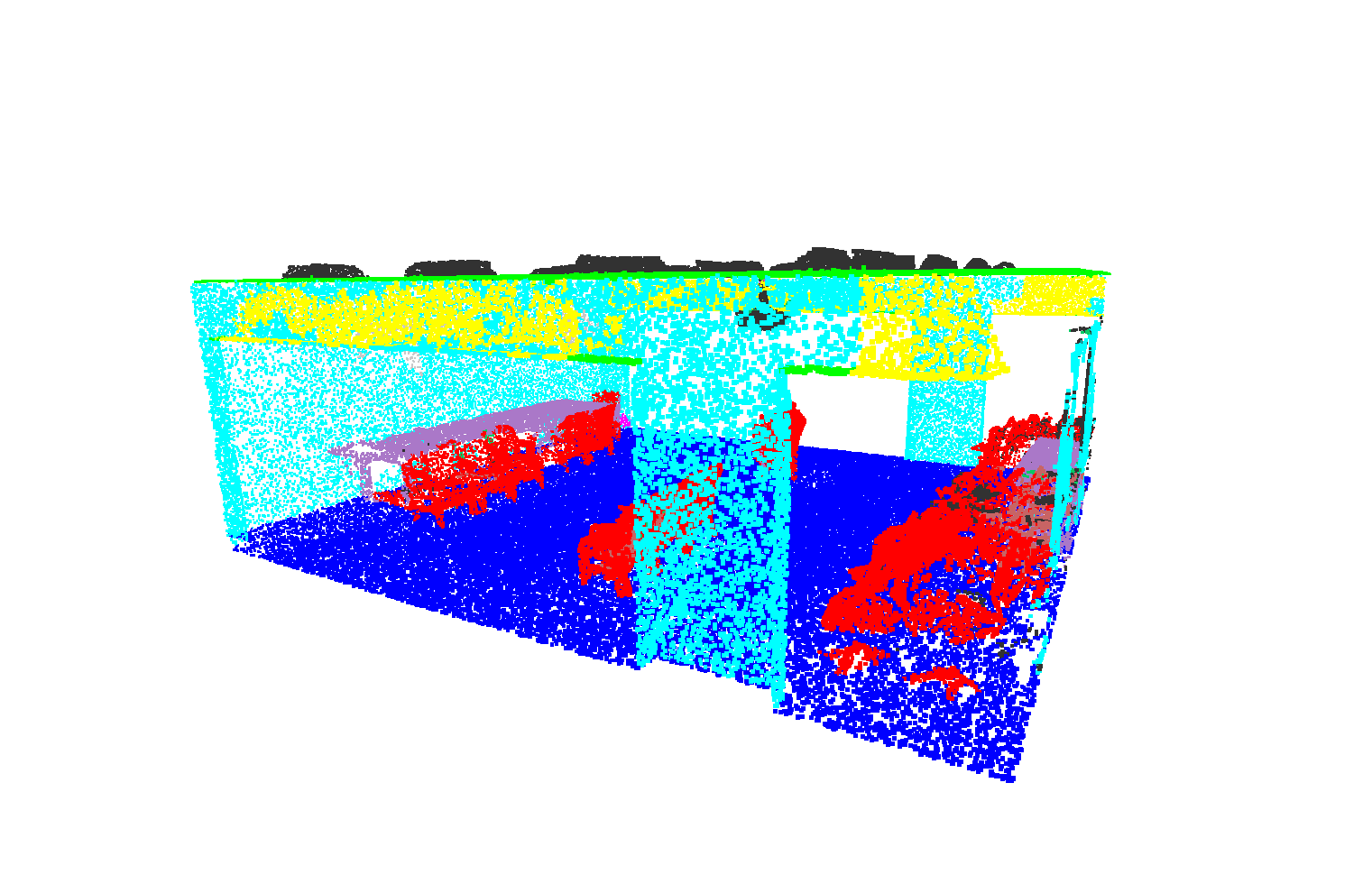}
    }
    \vspace{-3mm}
    \subfigure{
        \includegraphics[width=0.8\linewidth]{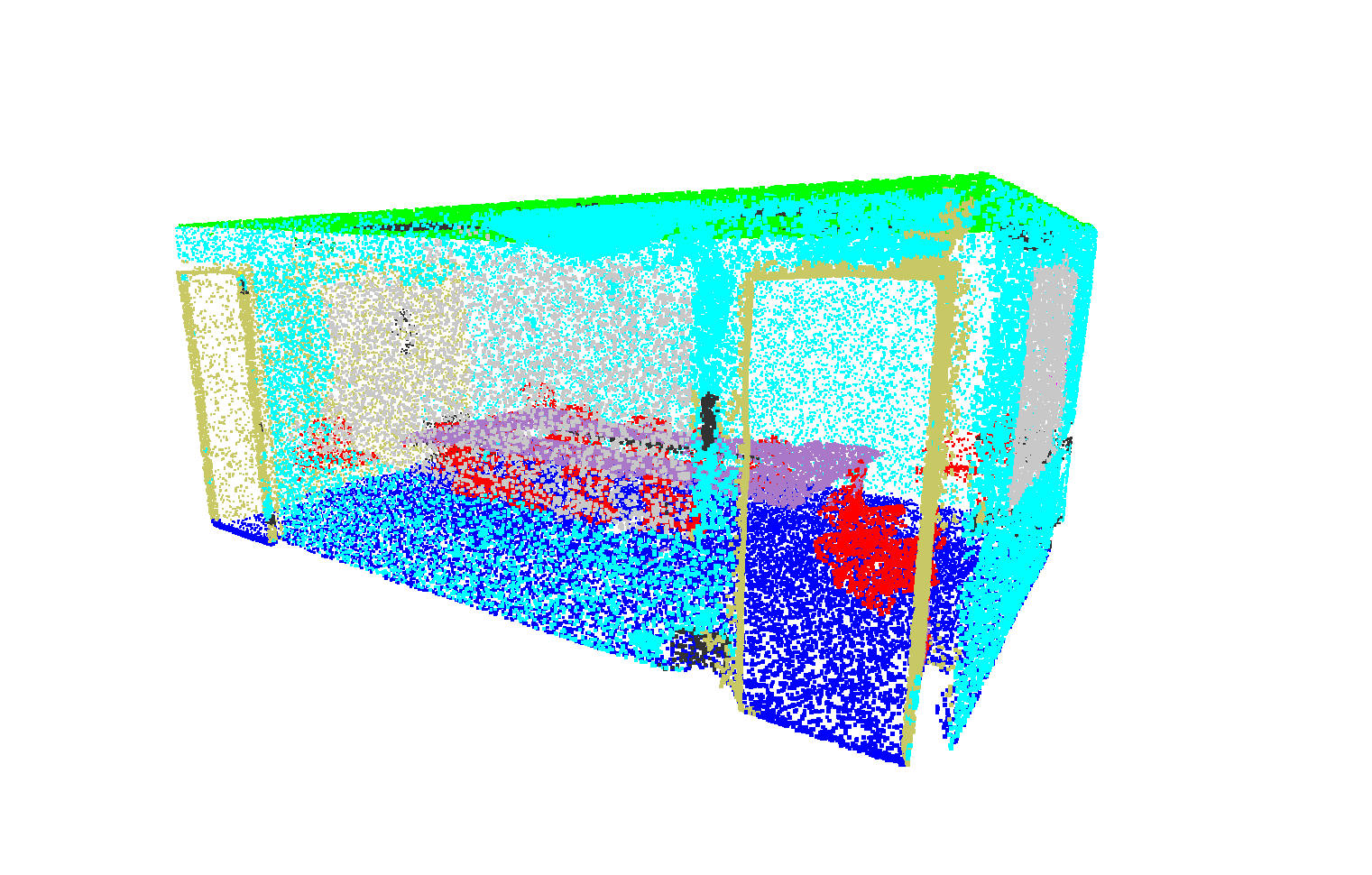}
    }
    \vspace{-2mm}
    \caption{More qualitative results for 3D semantic segmentation on S3DIS dataset~\cite{armeni20163d}.}
\end{figure}


{\small
\bibliographystyle{IEEEtran}
\bibliography{egbib}
}

\end{document}